EMPIRICAL STUDY

# Assessing the validity of new paradigmatic complexity measures as criterial features for proficiency in L2 writings in English

Cyriel Mallart[1], Andrew Simpkin[2], Paula Lissón[3], Nicolas Ballier[4], Rémi Venant[5], Bernardo Stearns[6], Jen-Yu Li[1], Thomas Gaillat[1]

[1]LIDILE, Université Rennes 2

[2]School of Mathematical and Statistical Sciences, University of Galway

[3]Universidad Internacional de la Rioja

[4]CLILLAC-ARP, Université Paris Cité

[5]LIUM, Université du Mans

[6]Insight, Data Science Institute, University of Galway

ABSTRACT
This article addresses Second Language (L2) writing development through an investigation of new grammatical and structural complexity metrics. We explore the paradigmatic production in learner English by linking language functions to specific grammatical paradigms. Using the EFCAMDAT as a gold standard and a corpus of French learners as an external test set, we employ a supervised learning framework to operationalise and evaluate seven microsystems. We show that learner levels are associated with the seven microsystems (MS). Using ordinal regression modelling for evaluation, the results show that all MS are significant but yield a low impact if taken individually. However, their influence is shown to be impactful if taken as a group. These microsystems and their measurement method suggest that it is possible to use them as part of broader-purpose CALL systems focused on proficiency assessment.

KEYWORDS
L2 English, proficiency assessment, microsystem; complexity measures, functional analysis, paradigm competitors

# 1. Introduction

Second Language (L2) writing assessment is an essential part of language education. It is a complex task which relies either on human judgement or computer-based measures. Since Page's first system (1968), assessing language writings with computers has generated much interest. Historically, such systems have always relied on automatic measurements, but they suffer from limitations in terms of construct representation (Cushing Weigle 2017). Addressing these limitations largely depends on the dimension of linguistic complexity.

Modern approaches to complexity are grounded in a mix of corpus linguistics and statistical methods of two types. Some metrics are holistic in nature and are used in proficiency predictive models (Lu 2014; Kyle 2016; Piln, Volodina, and Zesch 2016; Piln 2018), others target syntagmatic patterns mapped to specific functions as part of descriptive models (Biber and Gray 2011; Staples et al. 2016). Form-function networks (Ellis 1994, p. 377) reflect the functional organisation of a textual structure to create meaning. This meaning relies on the choice of forms rather than other possible options for the same context (Halliday and Matthiessen 2004, p. 24).

In both methodologies, tokens or lexico-grammatical patterns are identified and combined along the syntagmatic axis. When form-function mappings are operationalised, the assumption is that learners use patterns for one specific function, thus ignoring possible internal variations. Few studies have focused on analysing the internal structure of a form-function mapping. Yet, there seems to be paradigmatic instability within mappings, as learners show difficulties in linking some forms to specific functions. In fact, they seem to hesitate between several forms for the same function. We argue that it is important to understand what learners hesitate with when elaborating their discourse because hesitations are the cause of instability in L2.

A way to measure paradigmatic instability of the mappings is with *microsystems*. Instead of defining the language used as a system of independent units, it is considered as a set consisting of an undefined number of microsystems (Py 1980). When exposed to contexts, learners create microsystems of forms. In doing so they create original subsets that do not necessarily correspond to the canons of linguistic theory.

Analysing how these forms vary in relation to each other could cast new light on how learners tend to favour one form over the others, depending on context. Our assumption is that the forms' variations can be associated with developmental patterns in writings. Indeed, writings may include variations which could be centred around specific values and matched to specific developmental stages and proficiency levels. In a previous study, microsystems were analysed in terms of relative proportions of forms belonging to the same paradigm in learner texts (Gaillat et al. 2022). By adopting a supervised machine learning method, it was shown that such proportions were associated with proficiency. However, counting the actual use has some disadvantages. For example in the *proform* microsystem (it/this/that), using "that" compared to "it" or "this" was reported to be linked to higher CEFR levels. Hence, a new text using only "that" would receive the highest possible CEFR grade despite such a text being suboptimal. Another way of approaching the task, which circumvents this issue, is to consider the probability of occurrence of one form versus the others. In this respect statistical methods such as Generalized Linear Models (GLM), including logistic regression, provide solutions to calculate probabilities of using each form based on the local context.

Our new proposal is to characterise microsystems using the probability of one form relative to its counterparts within the same microsystem. To evaluate the discriminating potential of these microsystems in the measurement of proficiency, we assess how these probability distributions can be associated with proficiency. The rest of the paper is structured as follows: Section 2 lays out the theoretical background. Section 3 is dedicated to presenting our method. In Section 4 we present and analyse the results. Section 5 focuses on the discussion and Section 6 concludes on the findings and future perspectives.

## 2. Theoretical background

### 2.1. *Assessing writing ability and L2 proficiency*

The construct of writing ability is broad as it covers several underlying purposes. It is essential to distinguish between learning to write and writing to learn in areas other than language learning. Cushing Weigle (Cushing Weigle 2013) refines this distinction by identifying three purposes for assessment. Firstly, she considers Assessing Writing (AW) as a way to verify if students have skills in text production including revisions and pragmatic aspects. Secondly, she points out that Assessing Content through Writing (ACW) verifies whether students understand specific content. Finally, she defines the task at hand in this paper using Assessing Language through Writing (ALW). This task addresses whether students master "the second language skills necessary for achieving their rhetorical goals in English" (Cushing Weigle 2013) or not.

The purpose of ALW is to assign some level of proficiency to a written production in a foreign language. In order to establish an association between production and proficiency scientifically, some methods rely on alalytical rubrics (Knoch 2011), others on fine-grained checklists (Safari and Ahmadi 2023). In the case of Automatic Essay Scoring (AES), recent approaches of ALW have relied on neural network systems including the automatic extraction of non-explicit features. Conversely, traditional approaches to AES have relied on explicitly selected features that have been validated as being criterial for proficiency (Hawkins and Filipovi 2012).In this case, features may correspond to the operationalisation of linguistic complexity constructs such as lexical and grammatical diversity. As a result, linguistic complexity features and their measures are the foundation of many automatic proficiency assessment systems without which the writing ability construct may neither be measured, nor linguistically substantiated.

### 2.2. *Systemic grammatical complexity vs structural complexity*

Grammatical complexity has boasted a long tradition of research over the last few decades. With the advent of computer methods applied to corpora, textual measurement methods were applied to texts and a myriad of metrics were devised, not least the ubiquitous type-token ratio. Many of these measures were tested empirically on L2 texts, showing that, together with accuracy and fluency, learner variability in complexity could be used to evaluate language development. Measures belonging to the Complexity, Accuracy and Fluency (CAF) framework have appeared to provide objectivity in the way second language development is being analysed (Wolfe-Quintero, Inagaki, and Kim 1998, p. 3).

Bult and Housen's 2012 seminal work on complexity explored the linguistic dimensions actually covered by a large variety of measures. The authors proposed a taxonomy in which three components make up the construct, i.e., propositional complexity, discourse-interactional complexity and linguistic complexity. Our work relates to the third component more specifically interpreted as "a more stable property of the individual linguistic items", i.e. structure complexity (Bult and Housen 2012, p. 25). In their framework, structure complexity corresponds to distinct linguistic features considered at local level. The features' usage is analysed in terms of specific form counts or semantic function counts. Conversely, systemic complexity implies a context level analysis in which the diversity of different structures are accounted for. In both cases, the authors point out that any L2 analysis should accurately specify the type of complexity that is targeted, and what measure is used to operationalise the construct. Both types of linguistic complexity have been operationalised with many grammatical and lexical complexity measures.

Many studies have used complexity measures by operationalising lexical and grammatical complexity in a bid to model proficiency. In their survey of complexity measurement methods, Bult and Housen indicate that most studies include "general measures which tap global, overarching complexity constructs" (Bult and Housen 2012, p. 32). Such approaches rely on what Biber calls "omnibus" measures that "combine multiple structural and syntactic distinctions" in one value (Biber et al. 2020). These measures come with issues about the link between the grammatical or lexical constructs they are supposed to measure and what the measures actually account for. The use of the metrics is not always clear, thus leading to ambiguity in interpretations (Norris and Ortega 2008).

Albeit predictive, these measures fall short of explanatory power, which is a problem for linguistic approaches to the description of complexity, i.e. what do these multiplevariable measures tell us about the complexity of specific patterns used by learners? This is also a problem in the case of AES systems because such systems may use the features in formative assessment situations to provide explanations on the scores they contribute to predict, i.e. what linguistic features lead to a particular score? If the features are hard to intrepret linguistically, it makes it hard to convert its information into actionable recommendations. One way to address this problem of interpretability is to redirect efforts on measures that are clearly connected to linguistic functions and only target the structural complexity construct. By focusing on distinct structures, it is easier to clarify how linguistic features interact in the production of observed structures. Repetitive measurements of these features provide information on the evolution of structures across time or proficiency and thus inform on L2 developmental trajectories (Larsson et al. 2023; Staples et al. 2022).

Proponents of single structural complexity measurement rely on the registerfunctional approach in which specific linguistic patterns are associated with specific developmental stages (Biber and Gray 2011; Lan, Lucas, and Sun 2019; Biber et al. 2023; Biber, Larsson, and Hancock 2023; Larsson et al. 2023). By contrasting the use of clause structures versus phrase structures, these authors show that the development of L2 academic writing could be measured by singling out specific structures. In doing so, they mapped specific structural forms to specific syntactic functions and associated them with developmental stages. This follows the Variable

Competence Model (VCM) in which learners face and try to clarify competing rules and forms by removing non-standard forms and linking different forms to different functions (Ellis 1994, p. 376-377). This approach gives the benefit of model explainability because features reflect what the actualised forms in context mean. Nevertheless, it may only give a partial view of the form-function mapping paradigm. It may be argued that in the process of clarification, learners reach the optimal form-function mapping only gradually. There is variability in their arbitrary choices of forms when constructing meaning. They not only oppose form-function mappings with each other, but they oppose some forms with each other to express one function or meaning. Looking at the issue from Halliday's perspective, "meaning resides in systemic patterns of choice" (Halliday and Matthiessen 2004, p. 23), and systemic patterns correspond to the paradigmatic organisation of language. We argue that learners alternate forms on the paradigmatic axis to produce the same meaning, thus the same function.

For instance, nominal determination illustrates this issue with the fact that many learners tend to get confused between the use of *a*, *the* or article for the nominal determination function. The same phenomenon seems to happen with the dative alternation (Bresnan et al. 2007) where learners hesitate between the prepositional dative structure or the double object structure when they express the idea of giving something. Likewise, Gries and colleagues have developed a methodology (Gries et al. 2020; Gries 2022) to analyse constructions that compete with each other, e.g. the genitive against the N of N construction. So, form competition within one mapping also needs to be accounted for and neither current CAF measures, nor single form-function mappings address this need.

### 2.3.    *The microsystem as a component of structural complexity*

The notion of microsystems was first theorised in the 1980s as "a finalised system which is small enough" to be analysed and "large enough to give a clear idea of its finality" or its function (Gentilhomme 1980). This type of system is autonomous in the sense that its elements are grouped according to principles identified by the learners (Py 1980). While exposed to forms, the learner elaborates hypotheses regarding their relationships and their situation functions. A microsystem is a group of competing forms on the paradigmatic axis which are related by the same linguistic function.

The article microsystem composed of *a*, *the* or (zero article) illustrates this grouping principle because forms are mapped to the same function in spite of potential idiosyncrasy. For a description of , see for instance (Depraetere and Langford 2012). Examples (1), (2) and (3) show how *the* is used by learners of English in three samples from the EFCAMDAT corpus (Geertzen, Alexopoulou, and Korhonen 2013).

(1) "Ladies and Gentlemans, My flat was robbed the previous evening. In coming back at my home, I saw that *the* window was broken." (EFCAMDAT writing ID: 2498)

(2) * "What do you think about positive discrimination in *the* companies?" (EFCAMDAT writing ID: 569744)

(3) * "Why *the* gender's discrimination is still a problem in our society?" (EFCAMDAT writing ID: 579779)

In (1) the use of the article might be expected due to the associative anaphora linking *flat* and *window*. However, *the* is unexpected in (2) and (3) due to misunderstandings of the generic values of *companies* and *genders discrimination*. In examples (2) and (3), is in paradigmatic competition with *the* (Depraetere & Langford, 2012, pp. 9193). All three cases show that there is competition between the three forms when determining an entity. This competition and these idiosyncrasies result in instability. We do not know if this instability follows a random or a systemic pattern and, thus, if it is linked to L2 developmental patterns.

A microsystem is observed at local level and, because this construct relies on the functional organisation of the local structure, it can be categorised as a part of the structural complexity type (Bult and Housen 2012). In terms of operationalisation, many structural complexity indices have been created in the literature. They are based on counts of specific linguistic features linked to grammatical functions such as complement, adverbial or noun modifier clauses (Biber, Gray, and Poonpon 2011). These counts are accounted for per text by way of different aggregate methods such as mean and standard deviation (Wolfe-Quintero, Inagaki, and Kim 1998; Kyle 2016), but they do not consider potential competitor forms. A microsystem operationalises this form competition for the same function.

The construct of microsystems can be exploited in the analysis and assessment of L2 writing. Being structural makes it a good candidate to be considered in terms of frequency counts but also in terms of probability of occurrence. Its empirical validity calls for a number of requirements so as to use it for the description of writing development. As previously mentioned, it needs to be operationalised with a measurement method. This method must be evaluated in terms of how the measurements relate to actual L2 proficiency. Different patterns appear and disappear while L2 develops and these patterns can be associated to proficiency levels. One way of representing developmental stages is to use a proficiency scale such as the CEFR (Council of Europe 2018). This approach offers the advantage of linking learner development to a scale that is widely used by practitioners. Using the construct of microsystem raises three research questions:

(1) Which L2 microsystems could be mapped to single functions?
(2) How can these microsystems be measured?
(3) What is the validity of these microsystems regarding ALW measured in terms of proficiency?

The remaining sections of the paper are organised as follows: Section 3 outlines our method and describes our data. Section 4 delves into our results, and Section 5 provides a discussion of these results. Section 6 closes the paper.

## 3. Methods

In this section, we describe the corpora, the processing pipeline including the operationalised microsystems.

## 3.1. *Corpus data*

Two corpora are exploited in our experiments. The first one is the EFCAMDAT (Geertzen, Alexopoulou, and Korhonen 2013) in its refined version as described by Shatz 2020. The refined version of the EFCAMDAT is a collection of 723,282 writings collected online by *Englishtown* language schools across eleven countries. This version does not include C2 writings. The learners were required to write texts following prompts such as "introducing yourself by email" and "writing a movie review". Due to the variety of writing prompts, the types of genres were not controlled. *Englishtown* relies on language teachers to manually correct and grade the writings, which allows them to gradually move from one level to the next. It should be noted that inter-rater agreement was not reported regarding the consistency of grades between texts. In our study, the CEFR levels attributed to the texts actually correspond to the successful completion of coursework levels of *Englishtown* by these students. The completion of each *Englishtown* level is used as a proxy of their acquired skills. The writings span across 16 proficiency levels, which are mapped to the first five CEFR levels. Table 1 provides descriptive statistics of the distributions. In our study, the corpus is split between training and test sets for several modelling tasks applied to different microsystems.

The second corpus is used as an external validation dataset in order to evaluate the generalisation potential of our analyses. The *Corpus d'Etude des Langues Appliquées à une Spécialité* (CELVA.Sp) (Mallart et al. 2023) was collected in two French universities and includes 977 writings produced by graduate and post-graduate students (see Table 1). The writings were annotated by four language certification experts[1] who followed a protocol based on the descriptors of the writing production competence of the Common European Framework of Reference (CEFR) (Council of Europe 2018, Appendix 4, p. 187-189).

**Table 2.** Inter-rater agreement for a CEFR annotation task conducted by four raters on 30 writings sampled from the CELVA.Sp corpus

| Raters' pairwise agreement (Cohen's Kappa) | 1 | 2 | 3 | 4 |
|---|---|---|---|---|
| 1 | - | - | - | - |
| 2 | .52 | - | - | - |
| 3 | .79 | .61 | - | - |
| 4 | .76 | .55 | .75 | - |

To evaluate inter-rater reliability, we randomly and sequentially extracted two samples of 30 texts each that were annotated independently by the four annotators. An annotation-adjustment discussion session was conducted between the two samples. The Kappa results obtained for the first sample showed values ranging from .52 to .79. Table 2 shows the confusion matrix between the raters. The second sample showed also showed fair to good agreement as per Fleiss 2003, yet less than the 0.8 value

---
[1] These raters are teachers of English as a Foreign language with more than 20 years experience in the language

mentioned by Artstein & Poesio 2008. Permutation tests between the first and second sample results showed no significant difference (p-values of pairwise kappa differences >> .05). These results must be read in the light of the complexity of the task which involved classifying entire texts into one of five categories among four annotators. This clearly leads to a more difficult task for interpretation. Given the experience of the experts and the CEFR rubric that was used, agreement levels are acceptable. Each individual annotator was then given a split of the remainder of the corpus to annotate.

**Table 1.** Writings across CEFR levels in the EFCAMDAT and CELVA.Sp corpora.

| Writings | # of writings | | % of writings | | av # of words | | Standard Deviation | |
|---|---|---|---|---|---|---|---|---|
| CEFR | EFCAMDAT | CELVA.Sp | EFCAMDAT | CELVA.Sp | EFCAMDAT | CELVA.Sp | EFCAMDAT | CELVA.Sp |
| A1 | 341,155 | 90 | 47.16 | 8.61 | 39.07 | 141.43 | 14.41 | 83.16 |
| A2 | 215,344 | 324 | 29.77 | 31.00 | 64.53 | 206.40 | 17.91 | 99.03 |
| B1 | 116,539 | 358 | 16.11 | 34.26 | 94.75 | 261.16 | 21.47 | 124.59 |
| B2 | 40,238 | 212 | 5.56 | 20.29 | 134.86 | 319.93 | 33.22 | 141.44 |
| C1 | 10,006 | 53 | 1.38 | 5.07 | 169.34 | 398.70 | 26.59 | 161.94 |
| C2 | NA | 8 | NA | 0.77 | NA | 388.63 | NA | 152.04 |
| Total | 723,282 | 977 | 100 | 100 | 62.75 | 253.85 | 34.87 | 135.40 |

### 3.2. *Defining potential microsystems*

In a previous study, several microsystems were identified and tested in terms of proficiency level (Gaillat et al. 2022). In our study, we use seven of these microsystems for which we suspect potential acquisitional confusion because they are mapped to similar functions. They refer to specific syntactic or semantic functions and are described in Table 3. For instance, when refering to an entity or a whole clause, learners do understand the need to use a proform but they get confused between IT, THIS or THAT, leading to a potential semantic error. Learners may also get confused between relativizers, leading to a syntactico-semantic error.

### 3.3. *Operationalising the measurement of microsystems*

To operationalise the construct, we adopt a different approach from a previous study based on counts and proportions (Gaillat et al. 2022). As a microsystem represents text at local level in terms of constructions, it is possible to model its forms in terms centres of the two French universities. They have extensive experience in the national language certification examination (CLES) as well as other certifications such as the Cambridge and the TOEIC tests in higher education.

**Table 3.** The seven micro-systems considered for this study

| Microsystems | Components | Function | Examples of confusions |
|---|---|---|---|
| Proforms | it, this, that | reference to entity | The student cares for this/that/it |
| Multi-noun | compound, genitive, prepositional | Pairs of nouns functioning as compounds, genitive or prepositional phrase | She took a student loan/a student's loan/the loan of a student. |
| Articles | a, the or ∅ | determining a noun | a/the/∅ loan |
| Duration | for, since or during | complementing a verb with duration related information | The student has had this loan for/since/during 2 years. |
| Quantifier 1 | any, some | determining a quantity: one or more or unspecified respectively | Any/some students could help. |
| Quantifier 2 | many, much | determining an important quantity | Many/much hard-working students don't rest. |
| Relativiser | that, which, who | surbordinator refering to entity | The students who/that/which study. |

of probability of occurrence. In this paper, we propose a supervised learning method based on multinomial logistic regression as in (1). Each microsystem $Y_i$ for texts $i = 1,...,n$ can take any one of $K$ forms (e.g. proform can be it, that or this so $K = 3$), and is assumed to follow multinomial distribution with parameter $\pi_i = (\pi_{i1},...,\pi_{iK})$. In this framework, the probability of $Y$ taking some discrete value $k$ ($k = 1,...,K$) is modelled as a function of $P$ predictor variables $X_1,...,X_P$.

$$Pr(Y = k|X) = \frac{exp(\beta_{0k} + \beta_{1k}X_1 + \cdots + \beta_{nk}X_n)}{1 + exp(\beta_{0k} + \beta_{1k}X_1 + \cdots + \beta_{Pk}X_P)} \quad (1)$$

The model returns values indicating the predicted probability of using each form of a microsystem. The explanatory variables $X_1,...,X_P$ are the features extracted from the local syntactic context in which the forms appear.

### 3.4. *Feature extraction*

To extract the features, we process the data in two stages. First, we conduct automatic annotation of the entire corpus with the use of UDPipe (Straka, Haji, and Strakov 2016). UDPipe provides multilevel annotation relying on the Universal Dependency (UD) framework v2.0 (de Marneffe et al. 2021). Each token is associated to linguistic information ranging from Parts of Speech (POS) to UD and includes morphological features such as person and number.

Secondly, we developed a tool that identifies microsystem forms by way of Grew queries (Guillaume 2021). Grew is a tool relying on graph representations of sentences. The framework is based on nodes linked to each other by relations. It unifies constituent and dependency grammar analysis by considering all types of structures as graphs. Consequently, we designed queries relying both on Universal dependency relations and constituent analysis. For instance, we extract all occurrences of the quantifier *many* with the query exemplified in 2, where the QUANT pattern is the lemma MANY used as an adjective modifying the noun it precedes. This includes cases where there are adjectival and adverbial modifiers in between.

$$QUANT[lemma = \text{``many''}]; N[upos = NOUN]; N-[amod]->QUANT; \quad (2)$$

We defined patterns for all the forms of the seven microsystems. Applying them to texts yields an MS-specific dataset made up of the targeted patterns together with their linguistic features given by UDpipe. For instance, each POS[2], UD relation and morphological feature appears in dedicated feature columns. Note that not all morphological features apply to all types of tokens, resulting in cases where some morphological features are null. Likewise, some target tokens may be located near sentence boundaries leading to right or left context features being absent, thus null.

**3.5.** *Feature selection*

Prior to modelling, we conducted feature selection. First we removed features with more than 50% missing values among all texts, making allowances for features related to the beginning and ending of sentences. This was to avoid removing features where a microsystem occurs as the first or last word of a sentence. We also dropped UDpipe features that actually describe the forms such as *lemma*, *wordform* and *textform* as these would trivially explain the occurrence. Table 4 lists all the selected features for the proform microsystem (see Appendix A for the feature set of each microsystem).

---

[2] This includes UD POS and Penn Tree Bank (PTB) POS

**Table 4.** Features used for the proform microsystem model

| Feature type | Feature description |
|---|---|
| POS | Left context 3-gram POS (UD and PTB) |
| POS | Right context 5-gram POS (UD and PTB) |
| POS | POS of dependency head |
| Dependency | Head-dependency relation between form and head |
| Dependency | Normalised dependency distance to root |
| Tokens | Number of tokens in pattern TB CHECKED |
| Morphology | Next token number |
| Morphology | Next token person |
| Morphology | Next token mood (UD) |
| Morphology | Next token verb tense |
| Nationality | Nationality declared by learner |

### 3.6. *Classifying the forms*

To perform classification we employed multinomial logistic regression on the EFCAMDAT training data and predicted labels with a testing subset. We first randomly split the data into 80% training and 20% testing. The random sampling occurred within each class and preserved the overall distribution of the data. Secondly, as the training set was imbalanced in terms of forms (less forms of one type than its competitors), we randomly subsampled the set to the lowest number of the forms making up a given microsystem. For instance, as there are only 34,484 occurrences of THIS proforms in the corpus, we selected the same number of THAT and IT proforms at random. The test set was preserved as its imbalanced nature reflected the natural distribution of the forms.

To predict the forms in context, we fitted the multinomial regression model on the training data made up of MS forms (outcome variable) and their contextual features. We then applied it to the testing data to obtain probabilities of occurrence of each form in each slot.

### 3.7. *Evaluation method*

Evaluation required a four-fold process involving annotating microsystem data as a Gold Standard (GS), using this GS to evaluate the automatic extraction of MS, evaluating predictions of MS forms in context and evaluating the association between predictions and proficiency.

**Annotation** First, we prepared a GS to evaluate our tools against human annotation of MS. We used the external corpus (see Section 3.1 to create a subset for each microsystem. We randomly extracted seven subsamples of circa 250 occurrences, and three linguists[3] annotated whether the forms matched the definitions listed in the annotation guidelines previously prepared (See Appendix B for annotation details including guidelines). We computed inter-annotator agreement with the Fleiss' Kappa index (Fleiss, Levin, and Cho Paik 2003). The choice of this index was motivated by its ability to deal with categorical data and to correct for chance agreements (Larsson, Paquot, and Plonsky 2020). Disagreements were treated by the third linguist in charge of consolidation.

**Extraction** Based on the Gold Standard, we could then evaluate the extraction queries described in Section 3.4. We applied the queries to the same external corpus subsets as the GS and computed accuracy metrics (F1-Score, precision and recall). This yielded information on the quality of the extraction of the MS forms to see if we could apply them to extract MS forms from the entire training and tests sets for the classification task.

**Classification** The prediction performance of the microsystem models was evaluated at the word level on the test set of the EFCAMDAT corpus. We compared predicted forms with the actual forms used by learners. As the test sets were imbalanced, we chose to report balanced accuracy.

**Evaluating MS associations with proficiency** Finally, we performed two tests to analyse prediction distributions. A Kruskal-Wallis rank sum test allowed us to analyse differences between CEFR levels. We also used ordinal logistic regression to investigate whether an association between the predicted probabilities of microsystem use were associated with the odds of increasing the CEFR level. In the latter case, form probability distributions were aggregated at text level with the median. Odds ratios indicated potential effects of the increase of form usage on proficiency. The evaluation was conducted on the test sets of the EFCAMDAT and CELVA.Sp corpora.

**Table 5.** Inter-annotator agreement for each of the microsystems

| Microsystems | N | Fleiss Kappa | z | p-value |
|---|---|---|---|---|
| MS A THE ZERO | 160 | 0.938 | 19.36 | 0.000 |
| MS IT THIS THAT | 165 | 0.951 | 21.06 | 0.000 |
| MS MUCH MANY | 110 | 0.937 | 12.160 | 0.000 |
| MS MULTINOUN | 135 | 0.959 | 18.64 | 0.000 |
| MS SINCE FOR DURING | 165 | 0.886 | 19.545 | 0.000 |
| MS SOME ANY | 109 | 0.985 | 13.495 | 0.000 |
| MS WHICH WHO THAT | 165 | 1 | 22 | 0.000 |

---

[3] Two doctors in linguistics (the annotators) carried out the annotation, and a professor of linguistics (the consolidator) validated the cases in which differences were identified and corrected according to the annotation guidelines.

## 4. Results

### 4.1. *Microsystem annotation*

As a preliminary step to evaluate extraction, we built a GS to test how well candidate microsystem forms (described in Table 3) could be identified according to the linguistic functions they mapped. Human annotation of seven MS-specific subsets revealed a very high level of agreement between the two annotators (Table 5) including Fleiss' Kappa values mostly above 0.9. These results showed that mapping the forms to their linguistic functions casts little doubt among expert annotators. However, some differences remained, and these were treated individually. Apart from obvious issues due to cognitive tiredness, differences were due to ambiguities in learner language. Example 4 illustrates this issue with confusion around the use of THAT. The THAT proform annotation guidelines indicate: "Annotate THAT only as proform, not as determiner, nor adverbial, nor relativiser nor complementiser". However, in this example the context of occurrence was ambiguous as the learner did not insert the obligatory *it or that* leaving an ambiguity in interpreting the used *that* as either a proform in subject position or a complementiser. In these cases the consolidator advocated for not tagging the forms as proforms and for applying this to all similar cases in order to provide consistency. The annotation differences are listed in Appendix C.

(4) a. "My opinion in the invention on the web is *that* is allowed at the time to start to communicate more easily , to exchange document ."
   b. "Except *that* is also important to consider the negative outcomes we can get from it."

By treating all the differences individually, decisions were made according to each context by respecting coherence in their application as in the case for proform or complementiser *that*. This process resulted in a consolidated dataset used as GS in the remainder of the experiments.

### 4.2. *Microsystem extractions*

To evaluate feature extraction and more specifically the quality of the Grew queries, we applied the query tool to the GS made up of annotated sentences and identified formfunction mappings (see Section 3.4. Results show how well MS forms were extracted by the query tool (see Table 6). The F1-score appears to be above 0.7 for all MS, showing a satisfactory level of robustness for wide-scale extractions. This is confirmed by confusion matrices as in Figure 1 showing proform extractions.

**Table 6.** Quality of microsystem extractions in the GS dataset

| Microsystems | Support | Precision | Recall | F1-score |
|---|---|---|---|---|
| MS A THE ZERO | 160 | 0.77 | 0.79 | 0.77 |
| MS IT THIS THAT | 165 | 0.87 | 0.87 | 0.86 |
| MS MUCH MANY | 110 | 0.79 | 0.87 | 0.77 |
| MS MULTINOUN | 135 | 0.71 | 0.76 | 0.72 |
| MS SINCE FOR DURING | 165 | 0.83 | 0.83 | 0.82 |
| MS SOME ANY | 109 | 0.75 | 0.77 | 0.74 |
| MS WHICH WHO THAT | 165 | 0.88 | 0.88 | 0.87 |

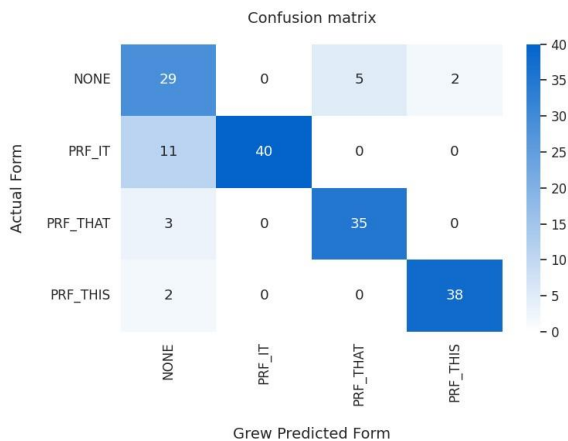

**Figure 1.** Confusion matrix for the extraction of IT, THIS and THAT proforms in the Gold Standard dataset

Precision and recall results show a good balance, indicating no strong bias towards either missing correct forms or badly identifying forms. Nevertheless, some issues remain regarding the A/THE/ microsystem. The Grew query does not capture the article very well (in front of nouns). Appendix D gives details about the extraction results, including itemised accuracy metrics of other MS forms.

### 4.3. *Classification of microsystem forms*

Knowing how well the query tool performed on the GS, we could apply queries to the entire training set and the EFCAMDAT test set. After conducting feature selection (see Section 3.5) we modelled the use of MS forms as an outcome using local context features as predictors. Applying multinomial logistic regression for classification, we obtained accuracy measures for each of the MS in the EFCAMDAT test set. Prediction performance of forms in context appeared to be a challenging task. Table 7 shows the consolidated results. For instance, we can see how well the local context can correctly predict the use of IT, THIS or THAT. For each proform, balanced accuracy shows that more than 70% of cases are predicted correctly by their local context features. Nevertheless, precision shows that THAT and THIS proforms tend to be mistagged. Similar results can be observed for other MS.

**Table 7.** Results for the MS classification with the EFCAMDAT test set

| Microsytems | Global accuracy (95% CI) | Balanced accuracy | Recall | Precision |
| --- | --- | --- | --- | --- |
| MS A THE ZERO | .91 (.9106, .9115) | .8752 / .8527 / .9976 | .8225 / .7406 / .9967 | .6689 / .8639 / .9990 |
| MS IT THIS THAT | .67 (.6675, .6731) | .7346 / .70163 / .71283 | .6924 / .563 / .583 | .932 / .291 / .240 |
| MS MUCH MANY | .87 (.8664, .8767) | .8695 | | |
| MS MULTINOUN {N2 N1/N1ofN2/N1sN2} | .56 (.5589, .563) | .6575 / .7029 / .70030 | .502 / .677 / .693 | .828 / .523 / .142 |
| MS SINCE FOR DURING | .73 (.7251, .7437) | .8228 / .7948 / .75147 | .7261 / .7432 / .681 | .5708 / .9407 / .3098 |
| MS SOME ANY | .82 (.8241, .8327) | .8169 | | |
| MS WHICH WHO THAT | .6 (.5979, .6143) | .7327 / .6704 / .7107 | .6559 / 05314 / .6376 | |

### 4.4. *Associations with CEFR levels*

This section reports the main findings of the study. Using the trained models for each MS, we obtained predictions for all the occurrences of the MS components in the internal and external test sets. Here, we investigated whether these predicted probabilities were associated with proficiency. We focus on the proform microsystem as an illustration. Figure 2 shows the variations of the median probabilities of each proform per text across the CEFR levels in the EFCAMDAT. The probabilities of IT seem to decrease as proficiency increases with significant differences between each level (Kruskal-Wallis rank sum test p-value <.01). THAT seems to be trending in the opposite direction, while THIS shows a slight variation for level A2 (Kruskal-Wallis rank sum test p-value <.01 in both cases). Plotting all MS shows unequal levels of variations depending on MS. The quantifier SOME ANY MS shows stark variations, while the relative pronoun MS reveals quite similar medians across CEFR levels of the EFCAMDAT.

Comparing probability distributions between the two test sets shows differences in several MS. This can again be illustrated with the proform MS (see Figure 3). Predictions between CEFR are not as clear-cut in the CELVA.Sp dataset. For instance, the Kruskal-Wallis rank sum test reveals that, while probabilities are significantly different in the EFCAMDAT between CEFR groups (N = 93,072, p < .001 for each proform), they are not in the CELVA.Sp (N = 905, p < .05 for THIS but p <.0.5 for IT and THAT). A close analysis of other MS reveals similar contrasting results, which suggests opposite trends in several cases. These differences might stem from the type of corpus data, including writing task and types of learners (see discussion in Section 5).

Finally, we performed ordinal logistic regression to investigate whether there was an association between the predicted probabilities of microsystem use and the odds of increasing CEFR level. Table 8 shows the results obtained for the EFCAMDAT set. Odds ratios indicate the odds of a better CEFR level for every 1% increase in use of one of the forms of aN MS. For instance, for a 1% increased probability of IT, the chances of improved proficiency drop slightly (.995 < 1). Conversely, for every unit increase of THAT, the chances of improved proficiency increase (1.011 > 1). In these cases, this would suggest that the probability of proform THAT vs IT and THIS tends to favour better proficiency. Similar observations can be made for MANY, N of N structures,

DURING and SINCE, SOME, and finally WHO and WHICH. All these forms tend to indicate better proficiency in the EFCAMDAT.

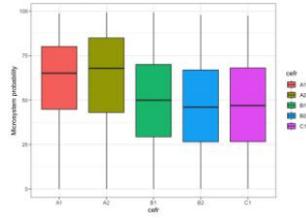

(a) Median probabilities of IT.

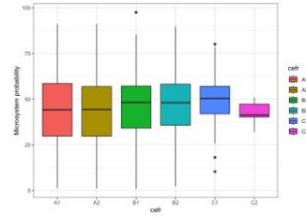

(a) Median probabilities of IT.

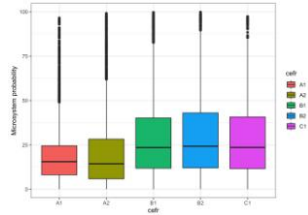

(b) Median probabilities of THAT.

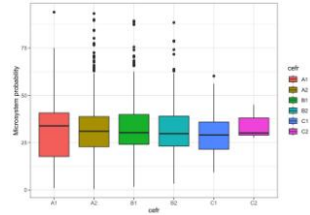

(b) Median probabilities of THAT.

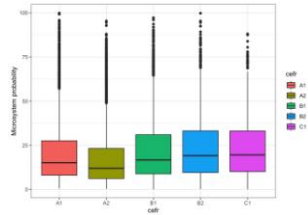

(c) Median probabilities of THIS.

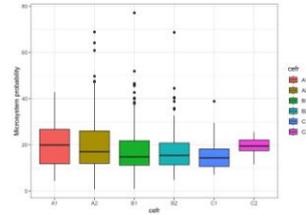

(c) Median probabilities of THIS.

**Figure 2.** Distribution of median probabilities of proforms per text in the EFCAMDAT test set.

**Figure 3.** Distribution of median probabilities of proforms per text in the CELVA.Sp external test set.

**Table 8.** Odds ratio between microsystems and CEFR levels in the EFCAMDAT and the CELVA.Sp.

| Microsystems | Components | EFCAMDAT Odds ratio | 95% CI | CELVA.Sp Odds ratio | 95% CI |
|---|---|---|---|---|---|
| MS DET | A | .998 | .997, .998 | 1.002 | .988, .1.015 |
|  | THE | 1.004 | 1.003, 1.004 | 1 | .987, .1.014 |
|  | 0 | **.996** | .996, .996 | **.999** | .996, 1.003 |
| MS PRF | IT | .995* | .995, .996 | 1.006* | 1, 1.013 |
|  | THAT | 1.011* | 1.011, 1.012 | .998 | .99, 1.006 |
|  | THIS | **.998*** | .997, .998 | **.974*** | .962, .986 |
| MS MLTNN | N2 N1 | **.993*** | .993, .993 | **.989*** | .976, 1.001 |
|  | N2 S N1 | **.999*** | .999, 1 | **.995** | .984, 1.005 |
|  | N1 OF N2 | **1.006*** | 1.006, 1.007 | **1.004** | .996, 1.013 |
| MS DUR | DURING | **1.005*** | 1.004, 1.007 | **1.004** | .995, 1.012 |
|  | SINCE | 1.008* | 1.007, 1.01 | 1 | .992, 1.007 |
|  | FOR | **.99*** | .989, .991 | **.991*** | .978, 1.004 |
| MS QUANT1 | ANY | .983* | .982, .983 | 1.004 | .998, 1.011 |
|  | SOME | 1.018* | 1.017, 1.018 | .996 | .989, 1.002 |
| MS QUANT2 | MANY | 1.012* | 1.011, 1.013 | .995 | .989, 1.001 |
|  | MUCH | .988* | .987, .989 | 1.005 | .999, 1.011 |
| MS REL | THAT | **.992*** | .99, .993 | **.979*** | .963, .995 |
|  | WHO | 1.001 | .999, 1.002 | .991* | .983, .999 |
|  | WHICH | **1.009*** | 1.007, 1.01 | **1.011*** | 1.003, 1.02 |

a * p-value <.05.

The same reasoning can be applied with the odds ratios obtained with the external test set. Some of the observations made on the EFCAMDAT remain the same, and they are shown in bold in Table 8. However, some other findings show opposite trends in the CELVA.Sp. We discuss this in Section 5.

We also conducted feature importance analysis with the varImp method in R's caret package (see Table 9 for details on the types of POS present in the right and left contexts of proforms). We computed the scaled importance, i.e. what percentage of the model each feature is responsible for. Important features vary according to the MS and their percentage distribution spreads considerably. For the proform microsystem, the largest feature percentage was 2.93% for the possessive in the right context of the the form with a 5 word window.

Overall, we can make some common comments regarding feature importance across all MS. Most important features were Penn-treebank POS tags as opposed to Universal POS tags also used in the models. This suggests that finer-grained morpho-syntactic annotation helps the classifiers. Finally, if we considered the top-10 features of each MS. Some MS appeared to rely on previous-context features (i.e., multinouns, quantifiers ANY/SOME,) while others mostly relied on post-context features (i.e., proforms). For others (i.e., quantifiers MANY/MUCH, relativizers and duration) both contexts were important.

**Table 9.** Top features used for the proform microsystem

|    | Feature        | Importance |
|----|----------------|------------|
| 1  | plus xposPOS   | 2.927      |
| 2  | plus xposNNS   | 1.757      |
| 3  | plus xposSYM   | 1.572      |
| 4  | plus xposFW    | 1.571      |
| 5  | plus xposEX    | 1.47       |
| 6  | plus xposJJS   | 1.425      |
| 7  | plus xposNN    | 1.383      |
| 8  | plus xposWRB   | 1.378      |
| 9  | minus xposRBR  | 1.304      |
| 10 | plus xposWP    | 1.281      |

## 5. Discussion

At the start of this paper, we raised three research questions in which we enquired about the relationships between microsystems and functions. Our experimental set-up was designed to operationalise, extract, predict and evaluate the predictions of MS forms in terms of proficiency. The purpose was to evaluate the use of probabilities of occurrence of MS forms as criterial features of proficiency. We first questioned the potential mapping between microsystems and meaning, leading to the identification of sets of forms to be chosen from in the same contexts. We showed that seven microsystems were mapped to specific functions including, proforms, quantifiers, relativisers, multinoun structures and articles. For each of these microsystems we operationalised their extractions with the use of consistent queries relying on multiple annotation layers in the corpora. These extractions were evaluated and results showed a very satisfactory performance.

The second question of the operationalisation of MS was central in our study. We used a novel approach to measure possible occurrences of forms. Rather than using proportions (as advocated in (Gaillat et al. 2022)), we used probabilities reflecting competition within each MS. Each model outputs a probability vector of its components and thus operationalises the concept of competition. The benefit of this approach is that instead of relying on the actual token used, the model actually relies on the contextual features that trigger the form. In fact, the model simulates what a learner would say for a specific slot of the context. Each MS model functions as an artificial learner specialised in certain grammatical constructions.

The validity of the construct was the final and crucial question of the study. The purpose was to evaluate whether the MS models' predictions could be linked to proficiency. Results included odds ratios showing the propensity of an MS to influence proficiency. A number of MS appear to be significantly associated with proficiency, albeit weakly. This raises the question of their actual importance for proficiency evaluation.

### 5.1. *Explaining CEFR variance with just MS?*

The percentage of CEFR variance explained with our seven micro-systems requires discussion since the odds ratios obtained in our results were statistically significant, but all close to 1 (suggesting lack of practical significance). We computed pseudo-$R^2$ to analyse how much of CEFR variance was actually captured by our regression models. The $R^2$ values were low (for the Proform MS, $R^2$ was 1.2% of the variation in CEFR). As a point of reference, Crossley and Kyle 2019 reported $R^2$ of 20% for a model made up of three cohesion variables. Our results showed MS odds ratios close to 1, but generally consistent across the internal and external test sets. We thus investigated further the practical significance of the MS construct.

To test this idea we ran a CEFR prediction model (multinomial regression LASSO) including all the MS as features with CEFR as the outcome variable in the test set of the EFCAMDAT. In this combined model we found $R^2 = 0.096$, meaning that 9.6% of the variation in CEFR was explained by the MS probabilities. This suggests that combining the MS together helped us to better understand CEFR.

In addition, Table 8 shows that some components of some MS are consistent across the two test sets. These components are more generalisable because despite differences regarding the diversity of tasks, the differences in L1 and other sociological differences between the two corpora, odds ratios remained similar. Another argument in favour of MS is that the MS markers are omnipresent in (learner) texts, and are more likely to account for CEFR variance than other tokens. What is the likelihood that any other word change in a text may trigger a change in the CEFR level? MS candidates correspond to low entropy sites[4], which indicates their tendency to accept few candidates per paradigmatic slot. As a result, their occurrence is more common than high entropy sites such as nouns or verbs.

### 5.2. *Causality vs association*

While we find evidence for associations between the probability of using a microsystem form and CEFR, this does not mean that increasing this probability will necessarily *cause* an increase in CEFR level. Assessing causality is not possible using these observational data. For instance, a learner in the EFCAMDAT wrote "That are a lot of computers. That are a lot of chars. That are a lot of desk. That are a lot of mouses. That are some flowers." This example shows that using THAT does not necessarily make a writing better. In other terms it, is not causal. This is important because even if an MS showed association, it would be risky to claim that students should write more of a particular component than the others in order to improve their level. Cross-examination with proficiency should be conducted in order to narrow down potential causality. Nevertheless, association is a first indicator of a potential issue. The advantage of the microsystem construct is that it points to a grammatical construct that can be easily interpreted. Associating an interpretable variable to proficiency helps understanding what makes a learner writing better or worse. This can be very helpful within the context of an Computer-Assisted Language Learning system focused on explaining proficiency classification.

---

[4] This can be tested with any Large Language Model, for example with the Hugging Face interface for BERT https://huggingface.co/google-bert/bert-base-uncased. A sentence like "If we know how to use it, [MASK] will improve our way of life" will output probabilities of occurrences much higher for the members of the proform MS: *it* (0.968), *this* (0.012), *that* (0.010) vs. *we* (0.003) or *they* (0.002).

### 5.3. *L1 effect*

Since the CELVA.Sp contains only French learners of English, we tested whether the results could be limited to an L1 effect with the proform MS. To do so we tested with/without L1 in the CEFR prediction model.

We tested this idea on the proform microsystem by introducing L1 as a variable in the ordinal logistic regression model. Results in odds ratios remained unchanged. For instance, odds ratio IT = .995 (without L1) when we obtained .996 (with the L1 included in the model). THAT has an odds ratio of 1.011 for both models. THIS has 0.998 and IT changes to .997. No confounding effect could be linked to the learner's L1, i.e. upon controlling for L1, the effect of our MS proform probability on CEFR remained the same.

### 5.4. *Task effect*

One important aspect to consider is that certain tasks in the corpus may elicit specific vocabulary and/or specific syntactic constructions. Previous research using the EFCAMDAT corpus has revealed effects of task type and instruction, with some tasks showing, for example, a higher number of pronouns (Alexopoulou et al. 2017), or a larger amount of complex noun phrases (Michel et al. 2019). Given the attested task effects in the corpus, it seems plausible that some tasks might be inadvertently eliciting a preference for one component of the MS. This might happen if one of the MS components is repeatedly used in the instructional prompt, making this component more salient than the others. Since each task is linked to a particular level, task effects might be a confounding factor in our study, because potential task effects affecting the MS would not appear consistently across the different proficiency levels.

It is worth noting that there are no predefined task categories in the EFCAMDAT corpus. Prior studies have established categories by inspecting the prompts and establishing similarities. For instance, Michel et al. (2019) propose the following task types: argumentation, description, instruction, narrative, comparison, and list/form. A full study of task effects in relation to the microsystems is beyond the scope of this paper, but we qualitatively explored part of the data to investigate whether task effects might be a confounding factor in our study.

We chose the MS formed by IT-THAT-THIS as an example, and plotted the normalised form frequency across topics of the EFCAMDAT by CEFR levels, which is shown in Figure 4. The distribution of THAT across CEFR shows a correlation with topic. This is an interesting case because in our model, the probability of THAT increases with higher proficiency. However, Figure 4 shows some outliers at the A1 level. These are defined as values of normalized frequency greater than 10. We examined the *EnglishTown* tasks represented by these outliers to analyse how much of our results could be explained by the instructional prompts. Out of the 14 outliers, 9 correspond to the same task, entitled *Taking inventory in the office* (Topic ID 2), in which students are asked to list the furniture and objects located in an office. While the students used THAT instead of there is/there are, we also see a lot of chars. That characterises the A1 level of EFCAMDAT writing. In one of the tasks, students were asked to describe an online shopping catalogue, and were producing sentences which, using images, might refer to images ("I like. This laptop is very nice, but that's is too expensive" EFCAMDAT writing ID 309259). It is also interesting that these three tasks target the same vocabulary. Overall, it seems that task effects might account for some of the outliers, but these are also related to misuses of components of the microsystems, particularly at low proficiency levels.

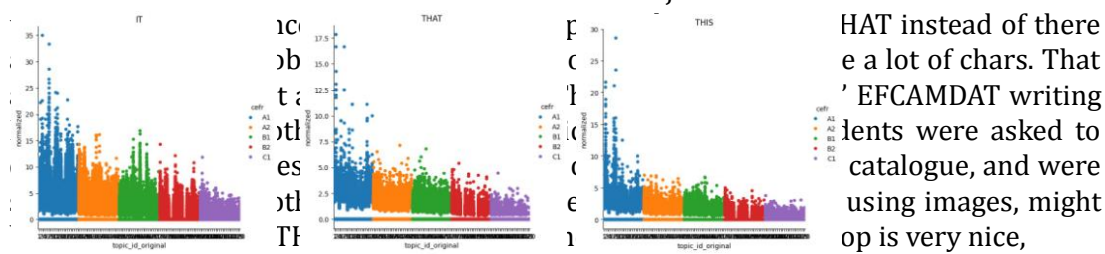

(a) Median probabilities of IT.   (b) Median probabilities of THAT.   (c) Median probabilities of THIS.

**Figure 4.** Normalized form frequency across topics of the EFCAMDAT.

### 5.5. *Designing learner trajectories*

Trajectory representation is a matter of using CEFR as an operationalisation of interlanguage stages. In this respect, MS probability distribution plots provide fine-grained views of the gradual changes, albeit subtle, that can be observed for each MS. Figure 2 illustrates this with the profoms, in which we see a small increase in the probability of use of THAT as CEFR increases. The MS construct can be seen as a method to measure and visualise learner trajectories with meaningful form-function mappings. This approach is similar to that of Biber and Staples's (2011) in which single form-function mappings are analysed as a function of proficiency levels and registers across the BAW corpus (Staples et al. 2022). Staples et al. examined the development of complexity features across university levels. Overall, they found that L1 writers used more phrasal features and fewer clausal features. Their results illustrated the learning trajectories, showing the variations in frequencies per feature type. O'Keeffe et al.'s project 2017 provided a mapping of lexico-functional patterns to CEFR levels via the use of multiple criteria including frequency, range of users and accuracy thresholds. In doing so, they created a map of patterns as a function of their onset across levels. The computations of patterns were based on occurrences.

In Saussure's representation of language, language (*langue*) is a system of systems. In our examples of MS, we focused on MS related to nominal structures and their determinations. Could we represent learner trajectories with such a limited linguistic scope? We acknowledge that the list of MS that we implemented is not exhaustive. We have started developing methods with other forms. Among the candidates that we consider are modals, but the UD annotation scheme does not allow queries for the epistemic/root distinction of modals. Defining new MS relies on the ability to design extraction queries that capture all element of an MS. Some MS may be very semantic which makes their extraction more difficult. As more tools are being developed for the characterisation of semantic and pragmatic features, it might become easier to extract the related MS.

## 6. Conclusion and perspectives

In this paper we reported on the findings of a study conducted on a novel set of structural complexity features. We theorised the existence of microsytems, i.e., groups of forms in paradigmatic competition when learners make their choices of words for specific linguistic functions. Instead of relying on frequency counts, we designed metrics indicating the probability of a form vs its competitors. Our purpose was to evaluate their utility with regard to proficiency.

We adopted a machine-learning approach which relied on the EFCAMDAT dataset to train seven microsystem models. These models were subsequently tested on an internal and an external test set. As the data were annotated with CEFR levels, we evaluated the associations between MS probabilities and CEFR levels. Results showed that all MS were significant but yielded low impact if taken individually. However, their influence was shown to be impactful if taken as a group. These microsystems and their measurement method suggest that it is possible to use them as part of broader-purpose CALL systems focused on proficiency assessment and feedback explanations.

Our approach appears to be precursory to language models in that it uses contexts for form predictions, but it focuses on restricted sets of linguistic forms. Language models provide probabilities based on the entire vocabulary set of their training corpus. Using large language models for the analysis of microsystems would be a logical next step. Their predictive power may provide finer results and pave the way towards the creation of models simulating artificial learners.

# 7. Appendices

## Appendix A. Tables of selected features for each microsystem prediction model

**Table A1.** Features used for the article microsystem model

| Feature type | Feature description |
| --- | --- |
| POS | Left context 5-gram POS (UD and PTB\textbackslash{}footnote{Penn Tree Bank tagset}) |
| POS | Right context 5-gram POS (UD and PTB) |
| POS | Head POS (UD and PTB) |
| Dependency | Head-dependency relation between form and head |
| Dependency | Normalized dependency distance to root |
| Tokens | Head token's position in sentence |
| Tokens | Position of MS token |
| Morphology | Number of head |
| Nationality | Nationality of the learner |

**Table A2.** Features used for the duration microsystem model

| Feature type | Feature description |
| --- | --- |
| POS | Left context 5-gram POS (UD and PTB\footnote{Penn Tree Bank tagset}) |
| POS | Right context 5-gram POS (UD and PTB) |
| Dependency | Head-dependency relation between form and head |
| Dependency | Normalized dependency distance to root |
| Tokens | Position of MS token |
| Morphology | token number in 2-gram right context and in 1-gram left context |
| Nationality | Nationality declared by learner |

**Table A3.** Features used for the quantifier any/some microsystem model

| Feature type | Feature description |
|---|---|
| POS | Left context 5-gram POS (UD and PTB\footnote{Penn Tree Bank tagset}) |
| POS | Right context 5-gram POS (UD and PTB) |
| POS | Head POS (UD and PTB) |
| Dependency | Head-dependency relation between form and head |
| Dependency | Normalized dependency distance to root |
| Tokens | Head token's position in sentence |
| Tokens | Position of MS token |
| Morphology | token number in 2-gram right context and in 1-gram left context |
| Nationality | Nationality of the learner |

**Table A4.** Features used for the quantifier many/much microsystem model

| Feature type | Feature description |
|---|---|
| POS | Left context 5-gram POS (UD and PTB\footnote{Penn Tree Bank tagset}) |
| POS | Right context 5-gram POS (UD and PTB) |
| POS | Head POS (UD and PTB) |
| Dependency | Head-dependency relation between form and head |
| Dependency | Normalized dependency distance to root |
| Tokens | Head token's position in sentence |
| Tokens | Position of MS token |
| Morphology | Token number in 1-gram right context |
| Nationality | Nationality of the learner |

**Table A5.** Features used for the multinoun microsystem model

| Feature type | Feature description |
|---|---|
| POS | Left context 5-gram POS (UD and PTB\textbackslash{}footnote{Penn Tree Bank tagset}) |
| POS | Right context 5-gram POS (UD and PTB) |
| POS | UPOS of the head of the dependency relation |
| Dependency | Head-dependency relation between form and head |
| Dependency | Normalized dependency distance to root |
| Nationality | Nationality of the learner |

**Table A6.** Features used for the relativizer microsystem model

| Feature type | Feature description |
|---|---|
| POS | Left context 5-gram POS (UD and PTB\footnote{Penn Tree Bank tagset}) |
| POS | Right context 5-gram POS (UD and PTB) |
| POS | Head's POS (UD and PTB) |
| Dependency | Head-dependency relation between form and head |
| Dependency | Normalized dependency distance to root |
| Tokens | Head token's position in sentence |
| Tokens | Position of MS token |
| Morphology | Mood of token in 1-gram right context |
| Morphology | Verb of token in 1-gram right context |
| Morphology | Tense of verb token in 1-gram right context |
| Morphology | Tense of head if verb |
| Morphology | Number in 1-gram left context |
| Nationality | Nationality of the learner |

## Appendix B. Annotation guidelines for the creation of the Gold Standard including microsystem forms.

Annotators were given a spreadsheet including one observation per line, i.e. an MS form. They were required to select the correct form among a list of possible candidates. The observations included proforms but also irrelevant forms used as disturbing variables.

**Annotation manual for microsystem patterns**

1. Open file with anotator's initials
2. Use the column annotation
3. Select an annotation cell
4. Read the sentence next to the cell and identify the place holder for the token toannotate. It is between two stars e.g., *this*
5. When selecting a cell, press alt key and arrow-down key to see the possible valuesto choose from. The none; value means the pattern does not correspond (because it has a different function in the context or because, for evaluation purposes, we have taken sentences that do not include the patterns).

Note: for ease of use and speed, it is advisable to use the keyboard keys. Table B1 lists the patterns and the definitions to comply with:

**Table B1.**

| MS | | Description |
|---|---|---|
| **Quantifier 1** | any | as a determiner |
| | some | as a determiner not as an adverbial |
| **Articles** | A/an | |
| | | Article A as a determiner |
| | THE | Article THE as a determiner |

| | Article 0 | Nouns without any determiner. As a proxy we list *nouns* that have neither determiner nor possessive pronoun dependency relation. In case there is a THE or A article in front of that noun, select the value corresponding to that article. If it is introduced by a quantifier (fewer, many, any...), select none. |
|---|---|---|
| **Proforms** | IT | It as an proform only, not extrapositional e.g. "it's ridiculous that they've given the job to PAt", nor impersonal e.g. "It seemed that / as if things would never get any better.". it-cleft constructions, e.g."It was your father who was driving - No it wasn't not, it was me." or weather/time it e.g. "It's only two weeks since she left.""It's raining." |
| | THIS | only as proform, not as determiner, nor adverbial |
| | THAT | only as proform, not as determiner, nor adverbial, nor relativizer nor complementizer. |
| **Multinoun** | | For the multinoun MS, the *last* word of the pattern is between two stars *. For instance: The university *car*; The university's *car*; The car of the *university* |
| | N of N | Any time a noun appears in a N of N construction |
| | NN | In cases of NN it can be either first or second position. e.g "I am studying materials science in an *engineering* school .". Here consider that the target to evaluate is Engineering school even if it is the first N that is between stars. NOTE: this pattern does NOT include ADJ + NN of course. |
| | N's N | Any time a noun appears in a N's N construction |
| **Duration MS** | | |
| | FOR | "For" used to express a lasting period of time (translates as "pendant" in FRench). Not to be confused with expression of purpose. e.g. "I want to do this for a gap year." or reason e.g. "thanks for doing xyz" |
| | SINCE | "Since" used as a point of departure in time |
| | DURING | "During" used for the expression of a lasting period of time |
| **Quantification** | | |
| | MUCH | Used to express quantity |
| | MANY | used to express quantity |
| **Relativizers** | | |
| | THAT | Uses of "that" as relative pronoun only, NOT as proform, determiner, complementizer or adverbial. |
| | WHICH | Uses of "which" as relative pronoun only, not as interrogative. NOTE: Watch relative pronouns as objects of verb. |
| | WHO | Uses of "who" as relative pronoun only, not as interrogative. NOTE: be careful with cases where WHO has no apparent antecedent: A who relative clause introduced by verb, e.g. "You can meet who you like" (Larreya & Rivire, 1991) |

# Appendix C. Differences in GS annotations between the two annotators and consolidation decisions

**Table C1.** Annotation differences for the proform microsystem in the Gold Standard

| Writing ID | sentence to annotate | annotation$_{TG}$ | annotation$_{PL}$ | consolidation |
|---|---|---|---|---|
| 5952 | Music makes me going through so much emotions and I love it *:* sadness , hapiness , nostalgia . | prf$_i$t | none | none |
| 1426 | My opinion in the invention on the web is *that* is allowed at the time to start to communicate more easily , to exchange document . | prf$_t$hat | none | none |
| 7552 | And *this* is it , at this very moment it stroke me . | prf$_t$his | prf$_t$his | prf$_t$his |
| 3129 | That 's why , I 'm contact you *,* because I need your help for my project . | prf$_t$hat | none | none |
| 3679 | I talked with people , teachers and students who have been or were still there , and I create in me a motivation about *this* project . | prf$_t$his | none | prf$_t$his |
| 5846 | Except *that* is also important to consider the negative outcomes we can get from it . | prf$_t$hat | none | none |

**Table C2.** Annotation differences for the quantifier some/any microsystem in the Gold Standard

| Writing ID | sentence to annotate | annotation$_{TG}$ | annotation$_{PL}$ | Consolidation |
|---|---|---|---|---|
| 814 | On social media , people shows what they think is nice to see , all the good in life but it transformes in bad because we don't see the reality in *any* of these photos . | quant$_a$ny | none | none |

**Table C3.** Annotation differences for the quantifier many/much microsystem in the Gold Standard

| Writing ID | sentence to annotate | annotation TG | annotation PL | consolidation |
|---|---|---|---|---|
| 6172 | The information is not filtering , he *many* have shocking photos or videos . | quant$_m$any | none | none |
| 8313 | Today we have *many* people than dislike the vaccine , and they do n't make it to their children . | quant$_m$any | quant$_m$uch | quant$_m$any |
| 10757 | The computer accumulate too *much* heat and the component melt . | quant$_m$any | quant$_m$uch | quant$_m$uch |
| 10950 | In world of mobile application developpement we have *many* tools which help to make an application . | quant$_m$uch | quant$_m$any | quant$_m$any |

**Table C4.** Annotation differences for the quantifier multinoun microsystem in the Gold Standard

**Table C5.** Annotation differences for the relativizer microsystem in the Gold Standard

| Writing ID | sentence to annotate | annotation TG | annotation PL | consolidation |
|---|---|---|---|---|
| 5 | Moreover this poem shows the *signification* of statue which is liberty , freedom , integration , american dream . | multinoun$_n$2$_{of_n}$1 none | | multinoun$_n$2$_{of_n}$1 |
| 10 | One of my favorite game is Outer Wilds , an exploration *game* where you play an archaeologist and an astronaut who travels across a tiny solar system in a spaceship to find clues about an antic civilisation. | multinoun$_n$2$_n$1 | multinoun$_n$2$_{of_n}$1 | multinoun$_n$2$_n$1 |
| 73 | I have the project to work in the *field* of cybersecurity . | none | multinoun$_n$2$_{of_n}$1 | multinoun$_n$2$_{of_n}$1 |
| 100 | Alex Dupont 2 : I choose the science of *education* for many reasons but not necessarily to become school teacher . | none | multinoun$_n$2$_{of_n}$1 | multinoun$_n$2$_{of_n}$1 |

**Table C5.** Annotation differences for the relativizer microsystem in the Gold Standard

| Writing ID | sentence to annotate | annotation TG | annotation PL | consolidation |
|---|---|---|---|---|
| No difference | | | | |

**Table C6.** Annotation differences for the duration microsystem in the Gold Standard

| Writing ID | sentence to annotate | annotation TG | annotation PL | consolidation |
|---|---|---|---|---|
| 4605 | Hello , my name is Alex Dupont and i am sending you this letter to explain to you my project *for* the next year . | none | dur$_f$or | none |
| 2654 | Moreover , this work experience is more importe *for* this year , it 's obligatory for a final evaluation . | none | dur$_f$or | dur$_f$or |
| 11225 | To finish i developed a critical spirit and *since* my redaction his better than before . | dur$_s$ince | none | none |
| 9012 | At the end of my second year of medicine , I worked *for* the first time in an EPHAD ( sort of center for elder person ) as an auxiliary . | none | dur$_f$or | none |
| 8723 | That 's why the laser surgery is important , because it gives people an other opportunity to correct their myopia , and this surgery improves your vision *for* life . | none | dur$_f$or | dur$_f$or |
| 9357 | By example , if a person sees a black cat *for* the first time , his eyes see it , send a message to the brain that connect its neurones and creates an engram relative to the black cat . | none | dur$_f$or | none |
| 5825 | I have a strong passion *for* reading and novels . | none | dur$_f$or | none |
| 6285 | People no longer have to wait *for* a specific time the news at the TV but they can research everythings at everytime on Internet . | dur$_f$or | none | dur$_f$or |
| 2337 | When the holidays arrived I passed all my time with her to help her to prepare her class *for* the following year . | none | dur$_f$or | none |
| 4362 | *Since* the first time i saw this Alex Dupont do his job with love and passion , i knew what i would like to do in my studies for the futur . | dur$_d$uring | dur$_s$ince | dur$_s$ince |
| 4204 | This project will be *for* the next year , and I want to go 6 months . | none | dur$_f$or | none |
| 4180 | Nevertheless , this remain my plan B *since* my plan | dur$_s$ince | none | dur$_s$ince |
| 9498 | I think , *for* the while , we ca n't stop the production of nuclear energy because we did n't find a energy enough efficient to substitute the nuclear energy . | none | dur$_f$or | dur$_f$or |
| 2029 | *Since* 3 years , I am interested by children . | dur$_d$uring | dur$_s$ince | dur$_s$ince |

## Appendix D. Quality of MS extractions in the GS dataset

**Table D1.** Quality of proform MS extractions in the GS

|  | precision | recall | f1-score | support |
|---|---|---|---|---|
| NONE | 0.64 | 0.81 | 0.72 | 36 |
| PRF IT | 1.00 | 0.78 | 0.88 | 51 |
| PRF THAT | 0.88 | 0.92 | 0.90 | 38 |
| PRF THIS | 0.95 | 0.95 | 0.95 | 40 |
| accuracy |  |  | 0.86 | 165 |
| macro avg | 0.87 | 0.87 | 0.86 | 165 |
| weighted avg | 0.88 | 0.86 | 0.86 | 165 |

**Table D2.** Quality of quantifier some/any MS extractions in the GS

|              | precision | recall | f1-score | support |
|---|---|---|---|---|
| NONE         | 0.38 | 0.69 | 0.49 | 16  |
| QUANT ANY    | 0.97 | 0.81 | 0.89 | 48  |
| QUANT SOME   | 0.90 | 0.80 | 0.85 | 45  |
| accuracy     |      |      | 0.79 | 109 |
| macro avg    | 0.75 | 0.77 | 0.74 | 109 |
| weighted avg | 0.86 | 0.79 | 0.81 | 109 |

**Table D3.** Quality of quantifier much/many MS extractions in the GS

|              | precision | recall | f1-score | support |
|---|---|---|---|---|
| NONE         | 0.37 | 1.00 | 0.54 | 11  |
| QUANT MANY   | 1.00 | 0.82 | 0.90 | 49  |
| QUANT MUCH   | 1.00 | 0.80 | 0.89 | 50  |
| accuracy     |      |      | 0.83 | 110 |
| macro avg    | 0.79 | 0.87 | 0.77 | 110 |
| weighted avg | 0.94 | 0.83 | 0.86 | 110 |

**Table D4.** Quality of relativizer MS extractions in the GS

|              | precision | recall | f1-score | support |
|---|---|---|---|---|
| NONE         | 0.62 | 0.88 | 0.73 | 32  |
| REL THAT     | 0.90 | 1.00 | 0.95 | 36  |
| REL WHICH    | 1.00 | 0.85 | 0.92 | 47  |
| REL WHO      | 1.00 | 0.80 | 0.89 | 50  |
| accuracy     |      |      | 0.87 | 165 |
| macro avg    | 0.88 | 0.88 | 0.87 | 165 |
| weighted avg | 0.90 | 0.87 | 0.88 | 165 |

**Table D5.** Quality of multinoun MS extractions in the GS

|                  | precision | recall | f1-score | support |
|---|---|---|---|---|
| MULTINOUN N2N1   | 0.66 | 0.85 | 0.74 | 27 |
| MULTINOUN N2OFN1 | 0.51 | 0.78 | 0.62 | 23 |
| MULTINOUN N2SN1  | 0.94 | 1.00 | 0.97 | 33 |
| NONE             | 0.73 | 0.42 | 0.54 | 52 |

|              | precision | recall | f1-score | support |
|--------------|-----------|--------|----------|---------|
| accuracy     |           |        | 0.71     | 135     |
| macro avg    | 0.71      | 0.76   | 0.72     | 135     |
| weighted avg | 0.73      | 0.71   | 0.70     | 135     |

**Table D6.** Quality of article a/the/zero MS extractions in the GS

|              | precision | recall | f1-score | support |
|--------------|-----------|--------|----------|---------|
| ART A        | 0.98      | 0.80   | 0.88     | 56      |
| ART NONE     | 0.61      | 0.88   | 0.72     | 25      |
| ART THE      | 1.00      | 0.91   | 0.95     | 56      |
| NONE         | 0.48      | 0.57   | 0.52     | 23      |
| accuracy     |           |        | 0.82     | 160     |
| macro avg    | 0.77      | 0.79   | 0.77     | 160     |
| weighted avg | 0.86      | 0.82   | 0.83     | 160     |